\pdfoutput=1

\documentclass[11pt]{article}

\usepackage{adjustbox}
\usepackage[final]{acl}
\usepackage{algorithm}
\usepackage{algpseudocode}
\usepackage{times}
\usepackage{siunitx} 
\usepackage{latexsym}
\usepackage{graphicx}
\usepackage{amsmath} 
\usepackage{booktabs}
\usepackage{hyperref}
\usepackage{amssymb} 
\usepackage{xcolor}
\usepackage{hyperref}       
\usepackage{url}            
\usepackage{amsfonts}       
\usepackage{nicefrac}       
\usepackage{array}
\usepackage{colortbl}
\usepackage{multirow}
\usepackage[table]{xcolor}
\usepackage{longtable}
\usepackage{todonotes}
\usepackage{xcolor}
\usepackage{todonotes}
\usepackage{makecell}
\usepackage{enumitem}
\usepackage{subcaption}
\usepackage[font=small,labelfont=bf]{caption}
\usepackage[T1]{fontenc}

\usepackage[utf8]{inputenc}

\usepackage{microtype}

\usepackage{inconsolata}

\usepackage{graphicx}
\usepackage[normalem]{ulem}  

\title{Are Two LLMs Better Than One?\\
A Student--Teacher Dual-Head LLMs Architecture for Pharmaceutical Content Optimization}

\author{
  \textbf{Suyash Mishra\textsuperscript{a}},  
  \textbf{Qiang Li\textsuperscript{b}},
  \textbf{Anubhav Girdhar\textsuperscript{c}}
\\ 
  \textsuperscript{a}Roche,
  \textsuperscript{b}Accenture,  
  \textsuperscript{c}Involead,
\\
  \small{
    \href{mailto:suyash.mishra@roche.com}{suyash.mishra@roche.com},
    \href{mailto:qiang.i.li@accenture.com}{qiang.i.li@accenture.com},
    \href{mailto:anubhav.girdhar@involead.com}{anubhav.girdhar@involead.com} 
    }
 }

\definecolor{rocheblue}{HTML}{005EB8}
\definecolor{rochelightblue}{HTML}{CCE5FF}
\definecolor{rochegray}{HTML}{F0F0F0}
\begin{document}
\maketitle


\begin{abstract}
Large Language Models (LLMs) are increasingly used for content creation in regulated domains such as pharmaceuticals, where content must be medical accurate and legally compliant. Traditional manual quality control (QC) is slow, error-prone, and create publication bottlenecks. To address this, we introduce a modular, LLM/VLMs-driven QC architecture, LRBTC (Language, Regulatory, Brand, Technical, Content Structure), implemented through a \textbf{Student–Teacher–HITL} architecture and \textbf{Waterfall} rule-filtering logic for scalable, verifiable content validation. This architecture is designed to provide verifiable, traceable, and scalable content optimization. Our methods achieve 83.0\% F1 and 97.5\% recall on \textbf{AIReg-Bench}, reducing missed violations five-fold compared with Gemini 2.5 Pro. On \textbf{CSpelling}, it improves mean accuracy by +26.7 \%.  Our error analysis further reveals that while current models are strong at detecting misspellings (92.5\% recall), they fail to identify complex grammatical (25.0\% recall) and punctuation (41.7\% recall) errors, highlighting a key area for future work. This work provides a practical, plug-and-play solution for reliable, transparent quality control of content in high-stakes, compliance-critical industries. \href{https://demo-of-content-optimizer-338849523617.us-west1.run.app}{Demo Access}.
\end{abstract}

\begin{figure*}[ht]
    \centering
    \vspace{-40pt}
    \begin{minipage}{0.9\textwidth}
        \centering
        \includegraphics[width=1\textwidth]{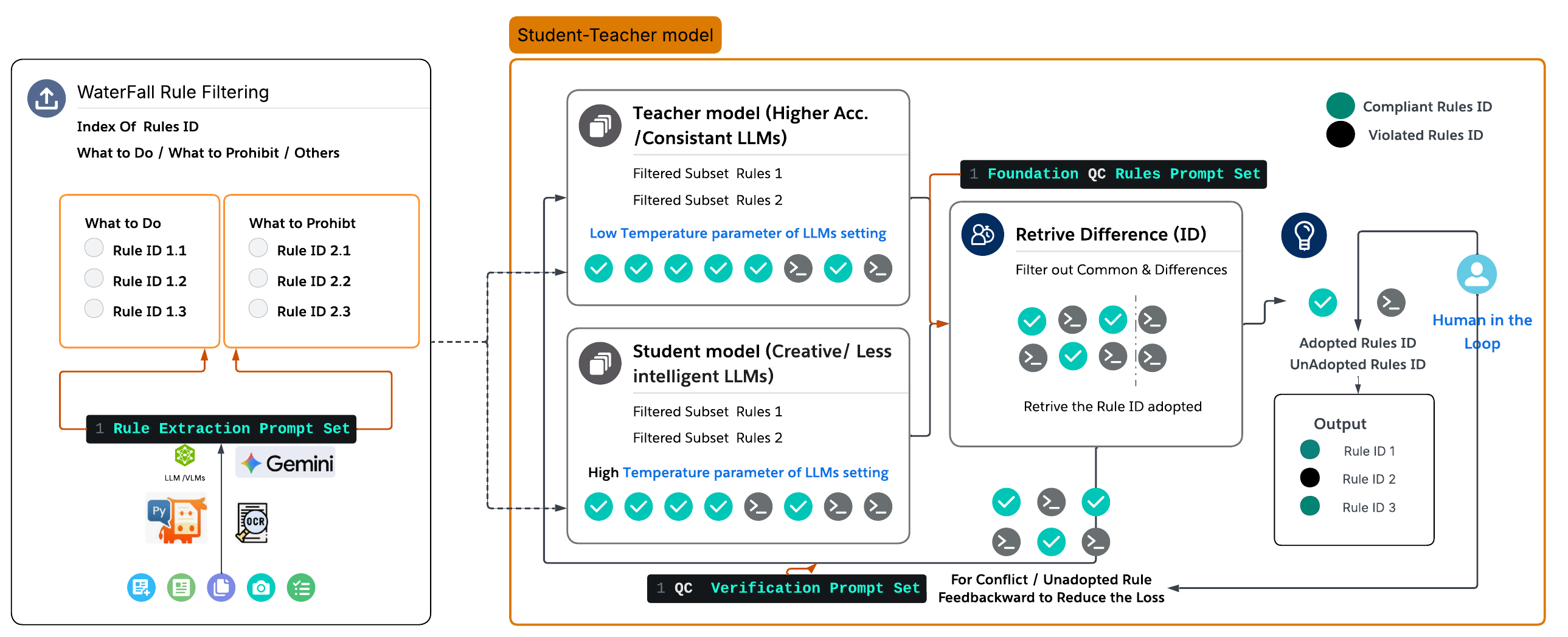} 
    \end{minipage}%
    \vspace{-5pt}
    \caption{System architecture of Student-teacher model to verify rule adoption. The teacher model guides the knowledge executed, and the student model verifies the commonly adopted rules, suggesting conflicts and new ideas. Iteratively verifies convergence into a common agreement, with a human in the loop to clean the leftover conflict rules. The core idea is that when knowledge is shared, common knowledge is enhanced and agreement is solidified. Conflicts or new ideas are often brought out by a counter-partner. Waterfall modeling can reduce the number of rules that need to be executed or checked. For rule filtering, we could apply a waterfall approach: IP - Country - Usecase - Topics - Subtasks (grammar/spell/etc), which would reduce and track the rules executed under each block.}
    \label{fig: architecture}
\end{figure*}

\vspace{-10pt}
\begin{figure*}[ht]
    \centering
    \vspace{-15pt}
    \begin{minipage}{0.9\textwidth}
        \centering
        \includegraphics[width=1\textwidth]{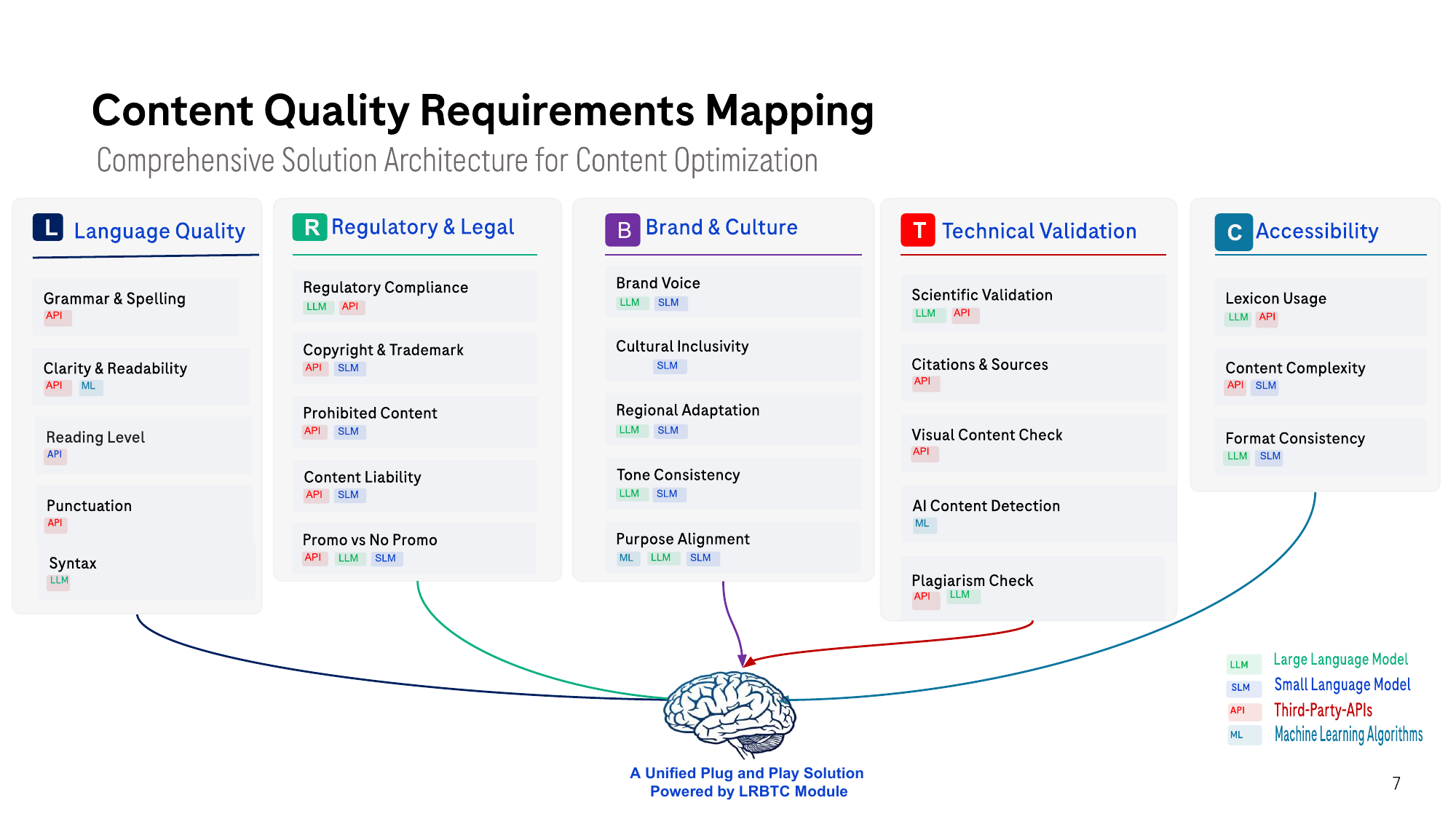} 
    \end{minipage}%
    \vspace{-10pt}
    \caption{Content Quality Requirements Mapping for Content Optimization.}
        \label{fig: architecture of LRBTC}
\end{figure*}

\section{Introduction}

Generating content with Large Language Models (LLMs) presents significant quality control (QC) challenges, particularly in regulated domains like pharmaceuticals. While LLMs accelerate content creation, their outputs can contain grammatical errors, factual inaccuracies, or formatting inconsistencies \citet{ji2023hallucination}, \citet{bang2023multitask}. In a sector like pharma, content must also be scientifically validated and compliant with strict legal and regulatory standards (e.g., FDA/EMA guidelines) \citet{fda2022labeling}.

Traditionally, this QC process is a multi-layered, manual review covering language, legal compliance, and scientific validity. This manual effort is time-consuming, prone to human error, and creates significant publication delays.

To address this bottleneck, we developed a modular Quality Control framework -- LRBTC, to automate and structure content validation. The LRBTC framework deconstructs the complex QC process into five clearly defined modules:

L (Language): Validates grammar, tone, clarity, and consistency.
R (Regulatory \& Legal): Checks for compliance with guidelines, prohibited words and legal liabilities.
B (Brand \& Culture): Ensures alignment with brand voice and cultural sensitivities.
T (Technical): Verifies technical and scientific elements, e.g, data, dosage information and citation accuracy.
C (Content Structure): Controls formatting, template adherence and reference.

In this paper, we demonstrate, via an GenAI NLP system, a platform that implements the LRBTC framework. Our system addresses the challenge of scaling this framework across diverse and large-scale data, including over 200,000 PDFs and 25,000 videos. (See appendix Table~\ref{table:roche-data})

Key challenges in implementing this framework are:
RQ-1: How to mine rules from which resource?
RQ-2: How to apply them across over 1000 rule sets without latency?
RQ-3: How do we verify them?

To solve this, our system, Optimizer (a plug-and-play QC platform) implements a student-teacher model combined with a waterfall logic and human-in-the-loop (HITL) verification. This approach allows us to distill complex rules from various sources (RQ-1) into efficient, deployable models (RQ-2) that provide verifiable and traceable outcomes (RQ-3). Our demonstration will showcase the LRBTC framework in action, applying it to both text-only and complex multimodal content. We demonstrate the system's effectiveness on leading benchmarks (e.g., Consumer Health CSpelling \citep{lu2019cspell} and AI RegBench \citep{guha2023legalbench}). Our contributions are:

\begin{itemize}
\item A modular, LLM-driven quality control architecture (LRBTC) for automated medical content validation.
\item A hybrid student–teacher and human-in-the-loop framework for  traceability.
\item Empirical evaluation on industrial multi-modal datasets and benchmark suites.
\end{itemize}

\section{Related Work}

\textbf{Spelling correction and medical text normalization.}  
\citet{lu2019cspell} introduced \textit{CSpelling}, a robust spell checker for consumer health queries, addressing non-word, real-word, and word-boundary errors with near real-time performance. 
\citet{kim2022cim} later proposed \textit{CIM}, combining character-level language models with corruption modules for context-sensitive misspelling correction, validated on CSpell and MIMIC-III. 
Such approaches remain essential for domain-specific normalization in clinical NLP. At a more general level, the \textit{CoLA} dataset \citep{warstadt2019cola} serves as a benchmark for sentence-level grammatical acceptability.

\textbf{Legal and regulatory compliance benchmarks.}  
\textit{AIReg-Bench} \citep{marino2025airegbench} provides the first dataset for assessing LLM compliance with the EU AI Act, featuring expert-labeled violation scenarios (e.g., Article~9) for reproducible evaluation. 
\textit{LegalBench Rule-QA} \citep{guha2023legalbench} similarly tests rule understanding and application in legal texts.

\textbf{Rule extraction and verification.}  
Recent research has progressed from heuristic prompts toward formal guardrails. 
\textit{Semantic Integrity Constraints (SICs)} \citep{lee2025sic} propose database-style abstractions for specifying and enforcing grounding and soundness constraints on LLM outputs. 
\textit{PROMPTEVALS} \citep{vir2025promptevals} collects developer-authored guardrails to assess coverage and constraint satisfaction. 


\textbf{Verification in multimodal models.}  
Recent benchmarks such as \textit{SCIVER} \citep{wang2025sciver} evaluate scientific claim verification in multimodal foundation models, with emphasis on factual grounding and citation consistency. Complementary studies such as HonestVQA \citep{anonymous2025the} and Decomposed NLI (DNLI) \citep{Yanuka_2025}, investigate calibrated self-verification, self-supervised reasoning, and hallucination auditing in long-context vision–language models \citep{gu2024mllmguard,li-etal-2025-llms}. These works also introduce new evaluation metrics, such as the Honesty Score (H-Score), Ethical Confidence Index (ECI), contradiction scores, and descriptiveness scores to quantify ethical calibration, factual consistency, and level of detail, as well as the degree to which generated captions contradict ground truth.


\section{Dataset And Experimental Settings}
\vspace{-10pt}
\begin{table}[ht]
\centering
\small
\caption{Agreement between LLMs and humans on AIReg-Bench.
Columns report quadratically weighted Cohen's scores $\kappa$, Spearman's $\rho$, Bias (measures the mean signed difference between each model's severity/violation score and the human-annotated ground truth, where values closer to 0 indicate better calibration), and MAE (lower is better).}
\vspace{-5pt}
\label{tab:llm-agreement}
\begin{adjustbox}{max width=\linewidth}
\begin{tabular}{@{}l
                S[table-format=1.3]
                S[table-format=1.3]
                S[table-format=-1.3]
                S[table-format=1.3]@{}}
\toprule
\textbf{Model} &
\textbf{$\kappa$ (↑)} &
\textbf{$\rho$ (↑)} &
\textbf{Bias (→0)} &
\textbf{MAE (↓)} \\

\midrule
GPT-5   \citep{openai2025gpt5}               & 0.849 & 0.838 & \bfseries -0.067 & \bfseries 0.450 \\
GPT-4o \citep{openai2024gpt4o}          & 0.775 & 0.842 & 0.458  & 0.558 \\
o3 \citep{openai2025o3}                  & 0.723 & 0.809 & -0.192 & 0.658 \\
o3 mini \citep{openai2025o3mini}                & 0.624 & 0.799 & 0.742  & 0.785 \\
Claude Sonnet 4 \citep{openai2025o3mini}    & 0.772 & 0.779 & -0.150 & 0.600 \\
\textbf{Gemini 2.5 Pro \citep{comanici2025gemini25}}& \bfseries \textbf{0.863} & \bfseries 0.856 & -0.225 & 0.458 \\
Gemini 2.5 Flash \citep{comanici2025gemini25flash} & 0.729 & 0.825 & -0.108 & 0.625 \\
Gemma 3 \citep{kamath2025gemma3}         & 0.696 & 0.757 & 0.258  & 0.692 \\
Grok 4 \citep{xai2025grok4}                   & 0.829 & 0.829 & 0.242  & 0.475 \\
Grok 3 mini \citep{xai2025grok3mini}             & 0.730 & 0.810 & 0.492  & 0.592 \\
\bottomrule
\end{tabular}
\end{adjustbox}
\end{table}

\vspace{-10pt}

\begin{table}[h!]
\centering
\caption{Model Performance Metrics on AIReg-Bench (N=120 systems) for compliance and EU AI Act rules, reported 95\% CI (Wilson score): [79.9\%, 93.9\%] on Gemini 2.5 Pro Recall, and [91.3\%, 99.3\%] for Our method Recall.} 
\label{tab:performance_comparison2}
\vspace{-5pt}
\begin{adjustbox}{max width=\linewidth}
\begin{tabular}{l c c}
\toprule
\textbf{Metric} & \textbf{Gemini 2.5 Pro (\%)} & \textbf{Our Methods (\%)} \\
\midrule
Accuracy ($\frac{TP+TN}{Total}$) & 65.9 & \textbf{75.9} \\
Recall ($\frac{TP}{TP+FN}$) & 88.7 & \textbf{97.5} \\
Precision ($\frac{TP}{TP+FP}$) & 65.1 & \textbf{72.1} \\
F1-Score  & 75.1 & \textbf{83.0} \\
Specificity ($\frac{TN}{TN+FP}$) & 34.4 & \textbf{43.8} \\
\bottomrule
\end{tabular}
\end{adjustbox}
\end{table}

\vspace{-10pt}
Here, we primarily benchmark 9 state-of-the-art Large Language Models (LLMs) across two complementary task dimensions: regulatory compliance validation on (AIReg-Bench)  and medical language quality control (grammar, punctuation, medical word spelling, and formality) on (CSpelling). To ensure robust comparison, we employed both correlation-based and classification-based evaluation metrics. Statistical consistency between model predictions and human annotations was quantified using quadratically weighted Cohen’s~$\kappa$, Spearman’s rank correlation coefficient~($\rho$), bias (mean signed difference, where values closer to $0$ indicate better calibration), and mean absolute error (MAE; lower is better). Unless noted otherwise, reported metrics reflect the automated system output prior to human review; Model-level classification performance was further assessed using standard metrics of Accuracy, Recall, Precision, F1-Score, Specificity,  Token count, Latency and API cost. 

\textbf{CSpell Medical Spelling Dataset.} \citep{lu2019cspell} provides diverse examples of misspellings and structural language errors collected from consumer health questions submitted to QA systems. It includes non-word and real-word misspellings, word-split and merge errors, and informal phrasing. This dataset enables testing of linguistic robustness in health-related contexts, as illustrated in Figure~\ref{fig: Cspelling}. 

\textbf{AIReg-Bench} \citep{marino2025airegbench} evaluates language models’ capability to assess regulatory compliance under the EU Artificial Intelligence Act (AIA) \citep{eu2024aia}. The AIA, effective since August~2024 \citep{lomas2024}, establishes harmonized requirements for AI systems placed on the EU market \citep{mahler2022risk}. High-risk AI systems are required to comply with obligations on risk management, data governance, record keeping, human oversight, and technical robustness (Articles~9,~10,~12,~14,~and~15). AIReg-Bench defines for each article ten representative use cases. Each use case contains two example software systems, and each system includes 2 \textit{violation} scenarios and 1 \textit{compliant} description, as illustrated in appendix Figure~\ref{fig: AIregbench}. 
These EU AI Act articles serve as ground-truth regulatory references for evaluating LLM performance across documentation quality, accuracy, robustness, and cybersecurity compliance, providing a realistic testbed for regulatory reasoning and claim validation. 


\section{Student–Teacher Model, Human-in-the-Loop (HITL), and Waterfall Approach}

\textbf{Student–Teacher Model.}  
We design a dual-LLM structure consisting of a \textit{teacher} and a \textit{student} model to verify and refine rule adoption. The teacher model guides knowledge execution and validation, while the student model re-evaluates adopted rules, identifying potential conflicts and proposing novel interpretations. The teacher model is characterized by higher factual accuracy and stability, whereas the student model favors creativity and broader exploration. For instance, we combine \textit{Gemini 2.5 Pro} (high-accuracy) as the teacher and \textit{Gemini 2.5 Flash} (faster but less accurate) as the student. Their temperature parameters are tuned accordingly: we set $\tau_T \approx 0.2$ while $\tau_S \approx 1.0 $ to encourage diverse outputs for all reported experiments. Detailed system-role instructions and verification instructions for both models are provided in the Appendix Table~\ref{tab:foundation_qc_rules}.

We also conduct ablation experiments by pairing different LLM families (e.g., \textit{Gemini 2.5 Pro} \citep{comanici2025gemini25} as teacher and \textit{Claude Sonnet 3.5} \citep{anthropic2025claude25sonnet} as student) to assess cross-model generalization (shown in Table \ref{tab:cspelling_ablation}, \ref{tab:performance_comparison ab}). The underlying principle is that knowledge is robust if a less smart but creative model can independently identify, agree, or challenge rule consistency proposed by a more capable one. Empirically, our results confirm that the student model often contributes creative refinements missed by the teacher.

\textbf{Human-in-the-Loop (HITL).}  
To prevent overreliance on LLM verification, we integrate human oversight into the conflict-resolution loop. In our demo interface, users can visualize discrepancies between the student and teacher outputs, including detected violations and rule IDs. Human experts can select or deselect flagged rules, add contextual justifications, and provide corrective feedback to the teacher model through dedicated verification loop. This feedback not only resolves conflicts but also updates the system’s knowledge with human-approved decisions, ensuring interpretability and accountability. Verification and feedback prompt instructions are detailed in the Appendix Table~\ref{tab:verification_prompt}.

\textbf{Waterfall Framework.}  
Given the scale and heterogeneity of enterprise rule bases, direct rule-by-rule verification is inefficient and error-prone. We therefore employ a hierarchical \textit{waterfall} filtering approach to progressively narrow the rule space. Rules are evaluated along the sequence: 
\textit{IP $\rightarrow$ Country $\rightarrow$ Use Case $\rightarrow$ Topic $\rightarrow$ Subtask (e.g., grammar, spelling, citation)}.  
This decomposition substantially reduces duplication and latency while mitigating hallucination risks and long-context memory limitations in LLMs.

For example, the pharm company global code base encompasses multiple regional and therapeutic layers—international (e.g., IFPMA\citep{ifpma2019code}, EFPIA, PhRMA, PAAB \citep{paab2021code}), regional (EU vs. US etc), and local therapeutic areas (22+). Input content types span six formats (emails, websites, promotional claims, etc.) and four major use-case modules (Foundation QC, Visual Control, Personalization, Compliance Guidance). Within the Foundation QC layer, as shown the \textbf{LRBTC} framework, covering approximately 30 spelling rules and 46 prohibited terms defined in the  Global Copy Style Guide (July 2025). To operationalize the waterfall, we employ LLM-assisted OCR and topic classification to extract rule books and compliance codes, automatically indexing each rule by section, category, and Index ID. Detailed rule-extraction instructions and examples are provided in the Appendix Table~\ref{tab:rule_extraction_prompt}.

\subsection{Ablation study: mathematical rationale of the Student–Teacher Dual Model Framework}

To isolate the contribution of each component, we evaluate four configurations: (1) teacher-only, (2) teacher+student, (3) +waterfall filtering, (4) full system pipeline+HITL. The Student–Teacher architecture in our Optimizer is inspired by the practical knowledge-distillation paradigm, where a stronger teacher model transfers stable knowledge to a weaker or more lightweight student model. Classical teacher-student knowledge distillation is often formulated as minimizing the divergence between teacher’s output distribution and student’s output distribution, commonly through a KL-divergence or cross-entropy objective \citep{hu2023teacherstudentarchitectureknowledgedistillation}. 

In our setting, however, the goal is not model compression but controlled content optimization under compliance constraints. Therefore, we reinterpret the student-teacher mechanism as a dual-model (Fig.\ref{fig: architecture}) verification process: 1) The teacher model acts as a stable, high-accuracy evaluator, guides knowledge execution.
2) The student model acts as a more exploratory evaluator that can challenge, rediscover, or expand the teacher’s decision, it verifies commonly adopted rules and proposes conflicts or new ideas.
3) The HITL layer resolves unresolved conflicts and the human reviewer resolves remaining disagreements (See our \href{https://demo-of-content-optimizer-338849523617.us-west1.run.app/}{Demo}).

\subsubsection{Teacher and Student Knowledge Verification}

Let the true but unknown violation label be:
\[
y_j \in \{0,1\},
where:
y_j = 1
\]
indicates that rule \(r_j\) is violated. The teacher model LLM produces an estimate:
\[
\hat{y}^{T}_j = f_T(x, r_j),
\]
and the student model do a separate verification against the rule and produces another estimate:
\[
\hat{y}^{S}_j = f_S(x, r_j).
\]

The teacher model \(f_T\) is configured for stable and conservative judgment. For example, it currently use a lower decoding temperature and and its prompt is designed to encourage stable, confident, and consistent outputs (see Table \ref{tab:verification_prompt}): $\tau_T \approx 0.2$.

The student model \(f_S\) is configured for broader exploration and alternative reasoning prompt setting, encouraging more creative and diverse responses. For example, it use a higher decoding temperature : $\tau_S \approx 1.0$.

Therefore, the teacher and student are not redundant replicas. They are two different estimators of the same latent compliance state \(y_j\). The teacher usually has lower variance and higher stability, while the student may have higher variance but better exploratory coverage. This difference is useful because the student may discover edge cases or alternative interpretations that the teacher misses, see Table \ref{tab:performance_comparison2} results: Recall and Precision.

\subsubsection{Why Two Models Can Reduce Missed Violations}

In compliance-critical domains, the most dangerous error type is the false negative. For a Student--Teacher system, Optimizer may flag a violation if either the teacher or the student detects it:
\[
\hat{y}^{ST}
=
\hat{y}^{T}
\lor
\hat{y}^{S}.
\]

Under this decision rule, the system misses a true violation only when both models miss it:
\[
P(FN_{ST})
=
P(\hat{y}^{T}=0
\land
\hat{y}^{S}=0
\mid
y=1).
\]

If the teacher and student make partially independent errors, this can be approximated as:
\[
P(FN_{ST})
\approx
P(FN_T)P(FN_S).
\]

Since probabilities are bounded between 0-1:
\[
P(FN_T)P(FN_S) < P(FN_T),
\]
provided that:
\[
P(FN_S) < 1.
\]
\vspace{-10pt}

This captures the key mathematical intuition behind the Student--Teacher design: \textbf{ it is less likely to classify a true violation as compliant}. The student does not need to outperform the teacher; it only needs to catch some violations the teacher misses. In this way, the second model contributes complementary recall, which is reflected in the results shown in Table~\ref{tab:performance_comparison2}, where the Student--Teacher method achieves a higher recall of \(97.5\%\).
The pure OR-rule predicts a recall/precision tradeoff; empirically we observe precision improving alongside recall. This is due to waterfall filtering, which removes irrelevant rules before either student and teacher model evaluates content, reducing the opportunity for spurious flags as well. 


\section{Main Results}

\vspace{-10pt}

\begin{table}[ht]
\centering
\caption{Performance fluctuation (accuracy) on the Consumer Health CSpelling dataset across grammar, punctuation, medical term spelling, and formality categories.}
\vspace{-5pt}
\vspace{-5pt}
\label{tab:fluctuation-corrected}
\resizebox{\linewidth}{!}{%
\begin{tabular}{c | c c | c c| c}
\toprule
\textbf{Sample Set}  & \textbf{Gemini 2.5 Pro} & \textbf{Gemini 2.5 Pro} & \textbf{Our} & \textbf{Our} & \textbf{$\Delta$} \\
\textbf{Size}  & \textbf{Mean (\%)} & \textbf{SD (\%)} & \textbf{ Mean (\%)} & \textbf{SD (\%)} & \textbf{(\%)} \\
\midrule
10 & 17.6 & 26.6 & 34.4 & 25.9 & +16.8 \\
12 & 13.5 & 18.3 & 43.7 & 27.0 & +30.1 \\
12 & 29.5 & 31.3 & 57.9 & 32.2 & +28.4 \\
12 & 21.2 & 23.5 & 48.0 & 28.2 & +26.8 \\
10 & 24.2 & 31.0 & 41.1 & 34.8 & +16.9 \\
12 & 13.0 & 17.4 & 44.9 & 27.0 & +31.9 \\
12 & 28.9 & 30.6 & 61.8 & 33.3 & +32.9 \\
\bottomrule
\end{tabular}%
}
\end{table}

\begin{figure}[ht]
    \centering
    \vspace{-20pt}
    \begin{minipage}{0.6\textwidth}
        \includegraphics[width=0.8\textwidth]{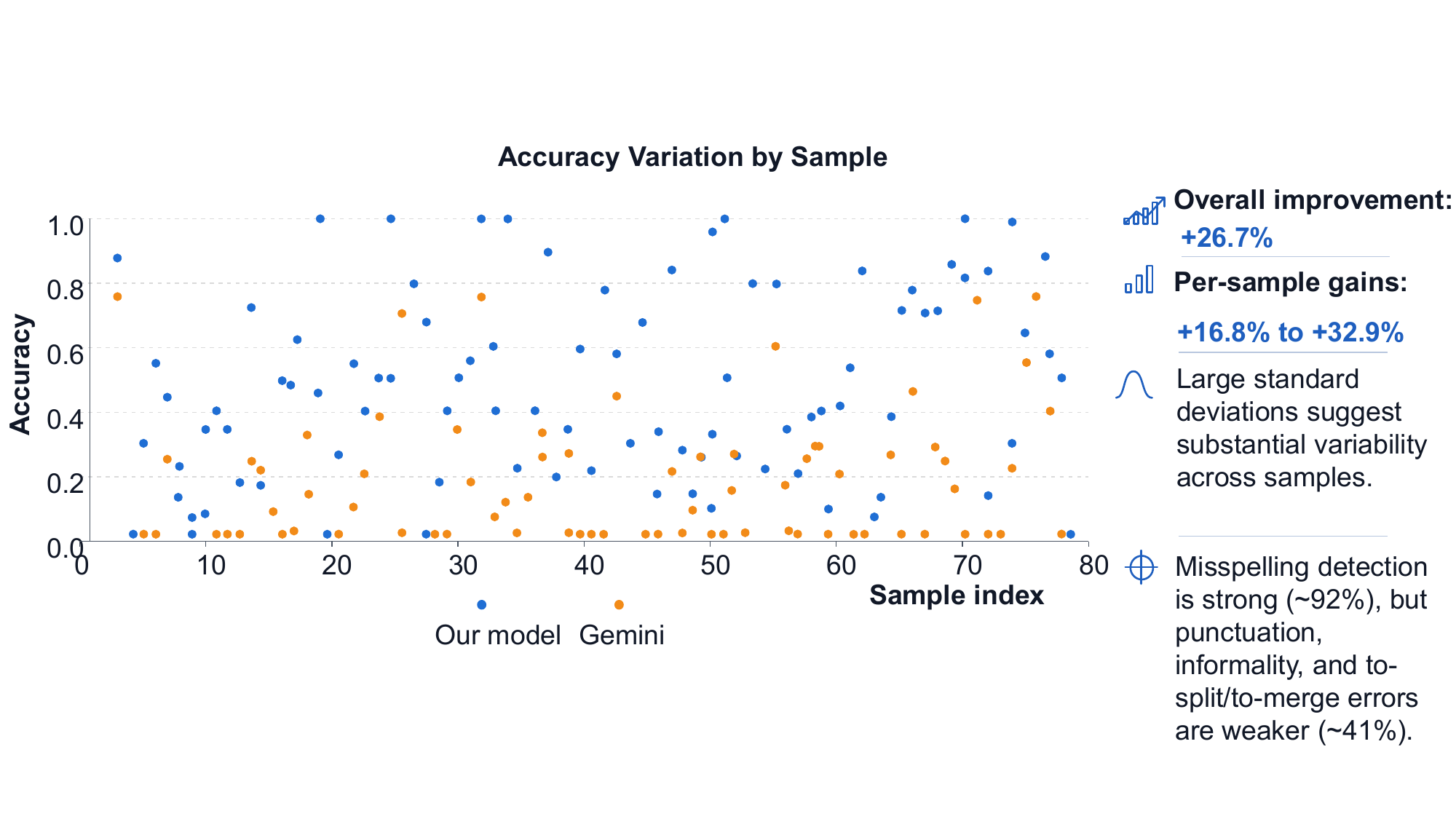} 
    \end{minipage}%
    \vspace{-10pt}
    \caption{Our model outperforms Gemini on all samples from Consumer Health Cspelling Dataset, with per gains ranging from  16.8\% to 32.9\% and overall improvement of 26.7\%. However, the relatively large standard deviations indicate substantial variability across sample, suggesting notable data heterogeneity. Both our and Gemini 2.5 pro are very good in detecting Misspelling error \textbf{(c.a. 92\%)}, but very bad on Punctuation, Informality, and To-split/To Merge errors \textbf{( c.a. 41\%)}.}
    \label{fig:cspelling-results}
\end{figure}

\begin{table}[h]
\centering
\caption{Error Detection Performance (Recall) by Failure Class on CSpelling}
\vspace{-5pt}
\label{tab:error-class-recall}
\resizebox{\linewidth}{!}{%
\begin{tabular}{l l l}
\toprule
\textbf{Error} & \textbf{\makecell{Total Occurrences}} & \textbf{\makecell{Detection}} \\
\textbf{Class} & \textbf{\makecell{(GT)}} & \textbf{\makecell{(Recall \%)}} \\
\midrule
Misspelling & {\(\approx\) 200} & 92.5 \\
ToSplit / ToMerge & {\(\approx\) 100} & 65.0 \\
Punctuation & {\(\approx\) 60} & 41.7 \\
Grammar & {\(\approx\) 100} & 25.0 \\
Informal / Word not exists & {\(\approx\) 40} & 12.5 \\
\bottomrule
\end{tabular}%
}\label{tab:errorclass}\vspace{-5pt}
\end{table}

\vspace{-5pt}

\begin{figure}[ht]
    \centering\vspace{-10pt}
    \begin{minipage}{0.5\textwidth}
        \centering
        
        \includegraphics[width=1\textwidth]{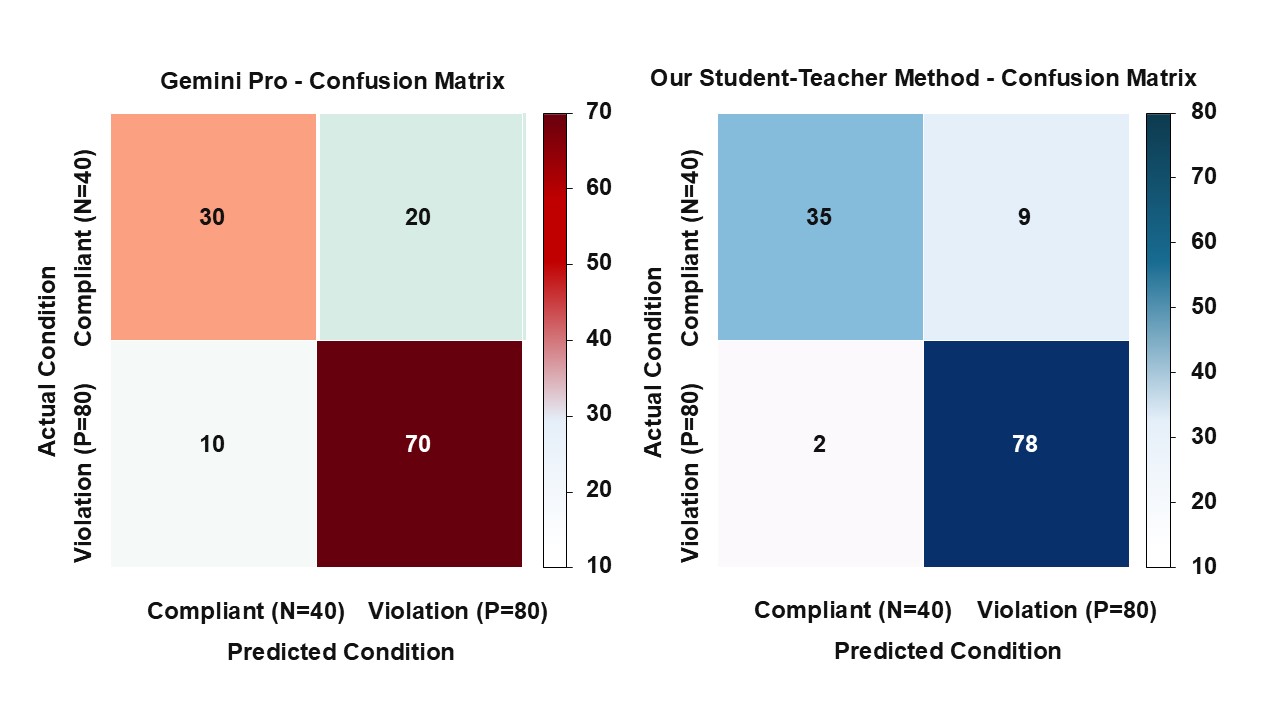} 
        \vspace{-20pt}
    \end{minipage}%
    
    \caption{Confusion Matrix Heatmaps on AIReg-Bench. These heatmaps visualize the raw counts of True Positives (Detected Violation), False Positives (Detect Compliance system into violation), True Negatives (Detect compliant system correctly), and False Negatives (Detect violation into compliant) across 120 EU AI Systems test cases. }
    \label{fig:confusionmatrix}\vspace{-5pt}
\end{figure}

\textbf{Regulatory Compliance Performance (AIReg-Bench)} Over 120 EU~AI~system test cases, our \textbf{Student–Teacher–HITL framework} consistently outperformed all baselines, shown in Table \ref{tab:performance_comparison2}. Compared with Gemini 2.5 Pro, our method achieved an absolute gain of \textbf{+10. in overall accuracy} (75.9 \% vs 65.9 \%), driven primarily by superior \textbf{recall} (\textbf{97.5 \% vs 88.7 \%}) and improved \textbf{specificity} (\textbf{43.8 \% vs 34.4 \%}). It is also important to note that our model's improvements are against a state-of-the-art baseline. As shown in \textbf{Table \ref{tab:llm-agreement}}, Gemini 2.5 Pro already achieved the highest human-agreement score ($\kappa=0.863$) among all 11 LLMs tested. The student–teacher verification loop proved especially effective for \textit{high-risk violation detection}: In a compliance context, failing to detect a violation (a False Negative) is the most critical error. The confusion matrix in Figure \ref{fig:confusionmatrix} shows our ST model correctly flagged 78 of 80 violations while maintaining balanced precision and low false-positive rates. In contrast, the Gemini 2.5 Pro baseline recorded \textbf{10 False Negatives}—a 5-fold increase in missed violations. This is directly reflected in the \textbf{Recall} score (the single most important metric for this task), where our model achieved \textbf{97.50\%} versus the baseline's 88.70\%. These confirm that iterative dual-LLM verification combined with human oversight improves both sensitivity and compliance reliability.

\textbf{Language Quality and Error Detection (CSpelling)}: Table \ref{tab:fluctuation-corrected} and Figure \ref{fig:cspelling-results} demonstrate consistent accuracy gains across all seven Consumer Health CSpelling subsets. Our approach yields an average improvement of \textbf{+26.7 \%} over Gemini 2.5 Pro, with per-set gains ranging from +16.8 \% to +32.9 \%. However, the relatively high standard deviations (25–33 \%) indicate substantial heterogeneity in data complexity. Combined with the erratic performance plotted in \textbf{Figure 3}, strongly suggests that performance is highly dependent on the specific data sample, indicating notable data heterogeneity. This suggests that STOA LLM future improvements should focus on syntactic and stylistic refinement rather than lexical correction alone.

\textbf{Error-Class Analysis and Variability} As shown in Table \ref{tab:errorclass}, recall varies dramatically by error class. Figure \ref{fig: Cspelling} further reveals that while both systems perform strongly on straightforward \textit{misspelling} cases ( 92 \% recall), performance drops sharply for \textit{punctuation} (41.7 \%), \textit{grammar} (25 \%), and \textit{informality} (12.5 \%). The Student–Teacher model’s architecture enables targeted gains in low-frequency and structurally complex errors (e.g., \textit{to-split / to-merge}, +24 pp recall over Gemini). Error analysis confirms that cascading rule filtering (Waterfall) reduces redundant checks and mitigates hallucination-induced misclassifications. 
Together, these results demonstrate that integrating structured rule extraction, student–teacher verification, and human oversight yields measurable, interpretable gains in both \textbf{regulatory compliance detection} and \textbf{medical-language quality control}.

\vspace{-5pt}

\section{Conclusion}
\vspace{-5pt}

In this work, we introduce an industrial framework for LLM-driven content optimization and compliance verification. 
Unlike prior studies that focus primarily on new model architectures, our contribution lies in proposing a \textit{plug-and-play Student–Teacher Model} designed to reduce LLM hallucination and enhance accuracy, consistency, and verifiability under realistic regulatory constraints. 
Our evaluation yields \noindent\textbf{Key Findings.} 
(1) \textbf{Compliance Robustness:} Our approach effectively enhances recall of high-risk violations while lowering false positives. 
(2) \textbf{Error-Type Asymmetry:} Misspelling detection is nearly saturated, but punctuation and grammar errors remain open challenges, calling for multimodal or syntactic fine-tuning. 
(3) \textbf{Rule Efficiency:} The Waterfall filtering mechanism reduces computational overhead by hierarchically pruning irrelevant rule sets, improving throughput without degrading accuracy.


\newpage



\newpage
\clearpage

\section{Limitations}

A main limitation of this study is to focus on justification of the domain-specific benefits, e.g. medical domain content optimization. We provide a baseline comparison of SOTA LLM/VLMs using both the AI-Reg benchmark and Consumer Health Cspelling dataset.  Currently, we focus on a real-world, high-stakes application domain (pharmaceutical compliance), where the need for reliable content generation is particularly strong. In future work, we will extend this line of research to other regulated domains, such as financial services and manufacturing \citep{li2022continuallearningdeploymentpipelines, 10217918, mishra-etal-2026-understanding, mishra-etal-2026-understanding}, to further validate the generalization of our solution blueprint.

\section{Acknowledgments}  

We sincerely thank Samik Adhikary and Puneet Srivastava for their sponsorship support from Roche. We also appreciate the insightful discussions and technical assistance provided by Qiang Li, Janina Kummerfeldt, Lynn Ma, Mayer Denis, and Kathrin Schwan from Accenture; Jennifer McGuire, Christopher Chu, Alex Connor, and Liz Stutz for business support from Roche; and Upender Phogat and Akshat Gupta for US MLP development support from Involead.

This Optimizer US-MLP use case, as well as the GenAI platform \citep{mishra-etal-2026-understanding} \citep{RICI}, would not have been possible without their contributions. We further extend our gratitude to the backend engineering teams who supported development, and to the healthcare professionals (HCPs), testers, and Roche Lab users whose consistent feedback brought our Optimizer use cases, particularly the Content Optimizer Agent to life and enabled continuous improvement.

This paper represents our academic contribution, in which we formalize experiments and evaluation methodologies using academic benchmarks. Through this work, we aim to share industry lessons learned and report valuable large-scale GenAI experiments in the pharmaceutical domain.

\bibliography{custom}

@misc{li2022continuallearningdeploymentpipelines,
      title={Continual learning on deployment pipelines for Machine Learning Systems}, 
      author={Qiang Li and Chongyu Zhang},
      year={2022},
      eprint={2212.02659},
      archivePrefix={arXiv},
      primaryClass={cs.LG},
      url={https://arxiv.org/abs/2212.02659}, 
}

@inproceedings{mishra-etal-2026-understanding,
    title = "From Understanding to Engagement: Personalized pharmacy Video Clips via Vision Language Models ({VLM}s)",
    author = "Mishra, Suyash  and
      Li, Qiang  and
      Girdhar, Anubhav  and
      Patil, Srikanth",
    editor = "Mysore, Sheshera  and
      Kumar, Sachin  and
      Balachandran, Vidhisha  and
      Hayati, Shirley Anugrah  and
      Brahman, Faeze  and
      Moussa, Hanane Nour  and
      Salemi, Alireza",
    booktitle = "Proceedings of the Second Workshop on Customizable {NLP}: Progress and Challenges in Customizing {NLP} for a Domain, Application, Group, or Individual ({C}ustom{NLP}4{U})",
    month = jul,
    year = "2026",
    address = "San Diego, California, USA",
    publisher = "Association for Computational Linguistics",
    url = "https://aclanthology.org/2026.customnlp4u-1.3/",
    doi = "10.18653/v1/2026.customnlp4u-1.3",
    pages = "24--43",
    ISBN = "979-8-89176-396-8"
    
}

@misc{hu2023teacherstudentarchitectureknowledgedistillation,
      title={Teacher-Student Architecture for Knowledge Distillation: A Survey}, 
      author={Chengming Hu and Xuan Li and Dan Liu and Haolun Wu and Xi Chen and Ju Wang and Xue Liu},
      year={2023},
      eprint={2308.04268},
      archivePrefix={arXiv},
      primaryClass={cs.LG},
      url={https://arxiv.org/abs/2308.04268}, 
}

@article{lu2019cspell,
    author = {Lu, Chris J and Aronson, Alan R and Shooshan, Sonya E and Demner-Fushman, Dina},
    title = {Spell checker for consumer language (CSpell)},
    journal = {Journal of the American Medical Informatics Association},
    volume = {26},
    number = {3},
    pages = {211-218},
    year = {2019},
    month = {01},
    issn = {1527-974X},
    doi = {10.1093/jamia/ocy171},
    url = {https://doi.org/10.1093/jamia/ocy171}
}

@misc{marino2025airegbench,
  author       = {Marino, Bill and Hunter, Rosco and Jamali, Zubair and Kalpakos, Marinos Emmanouil and Kashyap, Mudra and Hinton, Isaiah and Hanson, Alexa and Nazir, Maahum and Schnabl, Christoph and Steffek, Felix and Wen, Hongkai and Lane, Nicholas D.},
  title        = {{AIReg-Bench: Benchmarking Language Models That Assess AI Regulation Compliance}},
  howpublished = {arXiv preprint arXiv:2510.01474},
  year         = {2025},
  url          = {https://arxiv.org/abs/2510.01474}
}

@inproceedings{warstadt2019cola,
  author       = {Warstadt, Alex and Singh, Amanpreet and Bowman, Samuel R.},
  title        = {The Corpus of Linguistic Acceptability (CoLA)},
  booktitle    = {Proceedings of the Society for Computation in Linguistics (SCiL)},
  year         = {2019},
  pages        = {52--58},
  publisher    = {Association for Computational Linguistics},
  url          = {https://nyu-mll.github.io/CoLA/},
  note         = {Dataset and benchmark for grammatical acceptability judgments.}
}

@misc{gu2024mllmguard,
      title={MLLMGuard: A Multi-dimensional Safety Evaluation Suite for Multimodal Large Language Models}, 
      author={Tianle Gu and Zeyang Zhou and Kexin Huang and Dandan Liang and Yixu Wang and Haiquan Zhao and Yuanqi Yao and Xingge Qiao and Keqing Wang and Yujiu Yang and Yan Teng and Yu Qiao and Yingchun Wang},
      year={2024},
      eprint={2406.07594},
      archivePrefix={arXiv},
      primaryClass={cs.CL},
      url={https://arxiv.org/abs/2406.07594}, 
}

@inproceedings{guha2023legalbench,
  author       = {Guha, Neel and Nyarko, Julian and Ho, Daniel E. and Ré, Christopher and Chilton, Adam and Narayana, Aditya and Chohlas-Wood, Alex and Peters, Austin and Waldon, Brandon and Rockmore, Daniel N. and Zambrano, Diego and Talisman, Dmitry and Hoque, Enam and Surani, Faiz and Fagan, Frank and Sarfaty, Galit and Dickinson, Gregory M. and Porat, Haggai and Hegland, Jason and Wu, Jessica and Nudell, Joe and Niklaus, Joel and Nay, John and Choi, Jonathan H. and Tobia, Kevin and Hagan, Margaret and Ma, Megan and Livermore, Michael and Rasumov-Rahe, Nikon and Holzenberger, Nils and Kolt, Noam and Henderson, Peter and Rehaag, Sean and Goel, Sharad and Gao, Shang and Williams, Spencer and Gandhi, Sunny and Zur, Tom and Iyer, Varun and Li, Zehua},
  title        = {LegalBench: A Collaboratively Built Benchmark for Measuring Legal Reasoning in Large Language Models},
  booktitle    = {Proceedings of the 37th International Conference on Artificial Intelligence and Law (ICAIL ’23) – Datasets and Benchmarks Track},
  year         = {2023},
  url          = {https://arxiv.org/abs/2308.11462},
  note         = {Also available as arXiv preprint arXiv:2308.11462}
}

@article{lee2025sic,
  author       = {Lee, Alexander W. and Chan, Justin and Fu, Michael and Kim, Nicolas and Mehta, Akshay and Raghavan, Deepti and Çetintemel, Uğur},
  title        = {Semantic Integrity Constraints: Declarative Guardrails for AI-Augmented Data Processing Systems},
  journal      = {PVLDB},
  volume       = {18},
  number       = {11},
  pages        = {4073–4080},
  year         = {2025},
  doi          = {10.14778/3749646.3749677},
  url          = {https://arxiv.org/abs/2503.00600}
}

@inproceedings{vir2025promptevals,
  author       = {Vir, Reya and Shankar, Shreya and Chase, Harrison and Fu-Hinthorn, Will and Parameswaran, Aditya},
  title        = {PROMPTEVALS: A Dataset of Assertions and Guardrails for Custom Production Large Language Model Pipelines},
  booktitle    = {NAACL Long Papers},
  year         = {2025},
  url          = {https://arxiv.org/abs/2504.14738}
}

@article{wang2025sciver,
  author       = {Wang, Chengye and Shen, Yifei and Kuang, Zexi and Cohan, Arman and Zhao, Yilun},
  title        = {SciVer: Evaluating Foundation Models for Multimodal Scientific Claim Verification},
  journal      = {arXiv preprint arXiv:2506.15569},
  year         = {2025},
  url          = {https://arxiv.org/abs/2506.15569}
}

@InProceedings{kim2022cim,
  title = 	 {Context-Sensitive Spelling Correction of Clinical Text via Conditional Independence},
  author =       {Kim, Juyong and Weiss, Jeremy C and Ravikumar, Pradeep},
  booktitle = 	 {Proceedings of the Conference on Health, Inference, and Learning},
  pages = 	 {234--247},
  year = 	 {2022},
  editor = 	 {Flores, Gerardo and Chen, George H and Pollard, Tom and Ho, Joyce C and Naumann, Tristan},
  volume = 	 {174},
  series = 	 {Proceedings of Machine Learning Research},
  month = 	 {07--08 Apr},
  publisher =    {PMLR},
  url = 	 {https://proceedings.mlr.press/v174/kim22b.html}
  
}

@InProceedings{ji2023hallucination,
author = {Ji, Ziwei and Lee, Nayeon and Frieske, Rita and Yu, Tiezheng and Su, Dan and Xu, Yan and Ishii, Etsuko and Bang, Ye Jin and Madotto, Andrea and Fung, Pascale},
title = {Survey of Hallucination in Natural Language Generation},
year = {2023},
issue_date = {December 2023},
publisher = {Association for Computing Machinery},
address = {New York, NY, USA},
volume = {55},
number = {12},
issn = {0360-0300},
url = {https://doi.org/10.1145/3571730},
doi = {10.1145/3571730},
journal = {ACM Comput. Surv.},
month = mar,
articleno = {248},
numpages = {38}
}

@misc{fda2022labeling,
  title        = {Labeling for Prescription Drugs and/or Insulin: Guidance for Industry},
  author       = {{U.S. Food and Drug Administration}},
  year         = {2022},
  howpublished = {\url{https://www.fda.gov/}},
  note         = {Accessed: 2025-01-15}
}

@misc{anthropic2025claude25sonnet,
  title        = {Claude 2.5 Sonnet: Model Card and System Overview},
  author       = {{Anthropic}},
  year         = {2025},
  howpublished = {\url{https://www.anthropic.com}},
  note         = {Model Card / Release Website}
}

@misc{openai2025gpt5,
  title={GPT-5 System Card},
  author={{OpenAI}},
  year={2025},
  howpublished={\url{https://openai.com}},
  note={Accessed: 2025-01-15}
}

@misc{openai2024gpt4o,
  title={GPT-4o System Card},
  author={{OpenAI}},
  year={2024},
  howpublished={\url{https://openai.com}},
  note={Model Card}
}

@misc{openai2025o3,
  title={OpenAI o3 Model Card},
  author={{OpenAI}},
  year={2025},
  howpublished={\url{https://openai.com}},
  note={Release Notes}
}

@misc{openai2025o3mini,
  title={OpenAI o3-mini Model Card},
  author={{OpenAI}},
  year={2025},
  howpublished={\url{https://openai.com}},
  note={Model Card}
}

@misc{comanici2025gemini25,
      title={Gemini: A Family of Highly Capable Multimodal Models}, 
      author={Gemini Team and Rohan Anil and Sebastian Borgeaud and Jean-Baptiste Alayrac and Jiahui Yu  and et al.},
      year={2025},
      eprint={2312.11805},
      archivePrefix={arXiv},
      primaryClass={cs.CL},
      url={https://arxiv.org/abs/2312.11805}, 
}

@misc{comanici2025gemini25flash,
      title={Gemini: A Family of Highly Capable Multimodal Models}, 
      author={Gemini Team and Rohan Anil and Sebastian Borgeaud and Jean-Baptiste Alayrac and Jiahui Yu  and et al.},
      year={2025},
      eprint={2312.11805},
      archivePrefix={arXiv},
      primaryClass={cs.CL},
      url={https://arxiv.org/abs/2312.11805}, 
}

@misc{kamath2025gemma3,
      title={Gemma 3 Technical Report}, 
      author={Gemma Team and Aishwarya Kamath and Johan Ferret and Shreya Pathak and Nino Vieillard and Ramona Merhej and Sarah Perrin and Tatiana Matejovicova and et al.},
      year={2025},
      eprint={2503.19786},
      archivePrefix={arXiv},
      primaryClass={cs.CL},
      url={https://arxiv.org/abs/2503.19786}, 
}

@misc{xai2025grok4,
  title={Grok 4 Model Card},
  author={{xAI}},
  year={2025},
  howpublished={\url{https://x.ai}},
  note={Technical Documentation}
}

@misc{xai2025grok3mini,
  title={Grok 3 Mini Model Card},
  author={{xAI}},
  year={2025},
  howpublished={\url{https://x.ai}}
}

@misc{ifpma2019code,
  title        = {IFPMA Code of Practice},
  author       = {{International Federation of Pharmaceutical Manufacturers \& Associations}},
  year         = {2012},
  howpublished = {\url{https://www.ifpma.org/resource-centre/ifpma-code-of-practice/}}
}

@misc{paab2021code,
  title        = {PAAB Code of Advertising Acceptance},
  author       = {{Pharmaceutical Advertising Advisory Board}},
  year         = {2021},
  howpublished = {\url{https://code.paab.ca/}}
}

@misc{RICI,
      title={Scaling Vision Language Models for Pharmaceutical Long Form Video Reasoning on Industrial GenAI Platform}, 
      author={Suyash Mishra and Qiang Li and Srikanth Patil and Satyanarayan Pati and Baddu Narendra},
      year={2026},
      eprint={2601.04891},
      archivePrefix={arXiv},
      primaryClass={cs.CV},
      url={https://arxiv.org/abs/2601.04891}, 
}

@misc{explainableaibenchmark,
      title={XIMAGENET-12: An Explainable AI Benchmark Dataset for Model Robustness Evaluation}, 
      author={Qiang Li and Dan Zhang and Shengzhao Lei and Xun Zhao and Porawit Kamnoedboon and WeiWei Li and Junhao Dong and Shuyan Li},
      year={2024},
      eprint={2310.08182},
      archivePrefix={arXiv},
      primaryClass={cs.CV},
      url={https://arxiv.org/abs/2310.08182}, 
}

@misc{lomas2024,
  author       = {European Parliament},
  title        = {EU AI Act: first regulation on artificial intelligence},
  howpublished = {Website},
  year         = {2025},
  url          = {https://www.europarl.europa.eu/topics/en/article/20230601STO93804/eu-ai-act-first-regulation-on-artificial-intelligence}
}

@misc{eu2024aia,
  title        = {Regulation (EU) 2024/1689 of the European Parliament and of the Council of 13 June 2024 laying down harmonised rules on artificial intelligence and amending Regulations (EC)},
  author       = {{European Parliament and Council of the European Union}},
  year         = {2024},
  month        = jun,
  day          = {13},
  howpublished = {Official Journal of the European Union},
  url          = {https://eur-lex.europa.eu/eli/reg/2024/1689/oj/eng},
  note         = {Accessed 2026-02}
}

@misc{mahler2022risk,
  author       = {Mahler, Thomas},
  title        = {Between risk management and proportionality: The risk-based approach in the EU’s Artificial Intelligence Act Proposal },
  url           = {https://ssrn.com/abstract=4001444},
  year         = {2021}
}

@misc{anonymous2025the,
      title={The Confidence Paradox: Can LLM Know When It's Wrong}, 
      author={Sahil Tripathi and Md Tabrez Nafis and Imran Hussain and Jiechao Gao},
      year={2025},
      eprint={2506.23464},
      archivePrefix={arXiv},
      primaryClass={cs.AI},
      url={https://arxiv.org/abs/2506.23464}, 
}

@inproceedings{Yanuka_2025,
   title={Bridging the Visual Gap: Fine-Tuning Multimodal Models with Knowledge-Adapted Captions},
   url={http://dx.doi.org/10.18653/v1/2025.naacl-long.527},
   DOI={10.18653/v1/2025.naacl-long.527},
   booktitle={Proceedings of the 2025 Conference of the Nations of the Americas Chapter of the Association for Computational Linguistics: Human Language Technologies (Volume 1: Long Papers)},
   publisher={Association for Computational Linguistics},
   author={Yanuka, Moran and Ben-Kish, Assaf and Bitton, Yonatan and Szpektor, Idan and Giryes, Raja},
   year={2025},
   pages={10497–10518} }

@inproceedings{li-etal-2025-llms,
    title = "How {LLM}s React to Industrial Spatio-Temporal Data? Assessing Hallucination with a Novel Traffic Incident Benchmark Dataset",
    author = "Li, Qiang  and
      Tan, Mingkun  and
      Zhao, Xun  and
      Zhang, Dan  and
      Zhang, Daoan  and
      Lei, Shengzhao  and
      Chu, Anderson S.  and
      Li, Lujun  and
      Kamnoedboon, Porawit",
    editor = "Chen, Weizhu  and
      Yang, Yi  and
      Kachuee, Mohammad  and
      Fu, Xue-Yong",
    booktitle = "Proceedings of the 2025 Conference of the Nations of the Americas Chapter of the Association for Computational Linguistics: Human Language Technologies (Volume 3: Industry Track)",
    month = apr,
    year = "2025",
    publisher = "Association for Computational Linguistics",
    url = "https://aclanthology.org/2025.naacl-industry.4/",
    pages = "36--53"
   
}

@inproceedings{bang2023multitask,
      title={A Multitask, Multilingual, Multimodal Evaluation of ChatGPT on Reasoning, Hallucination, and Interactivity}, 
      author={Yejin Bang and Samuel Cahyawijaya and Nayeon Lee and Wenliang Dai and Dan Su and Bryan Wilie and Holy Lovenia and Ziwei Ji and Tiezheng Yu and Willy Chung and Quyet V. Do and Yan Xu and Pascale Fung},
      year={2023},
      eprint={2302.04023},
      archivePrefix={arXiv},
      primaryClass={cs.CL},
      url={https://arxiv.org/abs/2302.04023}
}

@article{1,
  title={Active retrieval augmented generation},
  author={Jiang, Zhengbao and others},
  journal={arXiv preprint arXiv:2305.06983},
  year={2023}
}

@article{2,
  title={Dense passage retrieval for open-domain question answering},
  author={Karpukhin, Vladimir and others},
  journal={arXiv preprint arXiv:2004.04906},
  year={2020}
}

@INPROCEEDINGS{10217918,
  author={Ozmermer, Evrim and Li, Qiang},
  booktitle={2023 IEEE 21st International Conference on Industrial Informatics (INDIN)}, 
  title={Self-supervised Learning with Temporary Exact Solutions: Linear Projection}, 
  year={2023},
  volume={},
  number={},
  pages={1-7},
  doi={10.1109/INDIN51400.2023.10217918}}

\appendix
\clearpage
\newpage
\section{Appendix}
\label{sec:appendix}

\begin{figure*}[t]
    \centering
    \begin{minipage}{1\textwidth}
        \centering
        \includegraphics[width=1.1\textwidth]{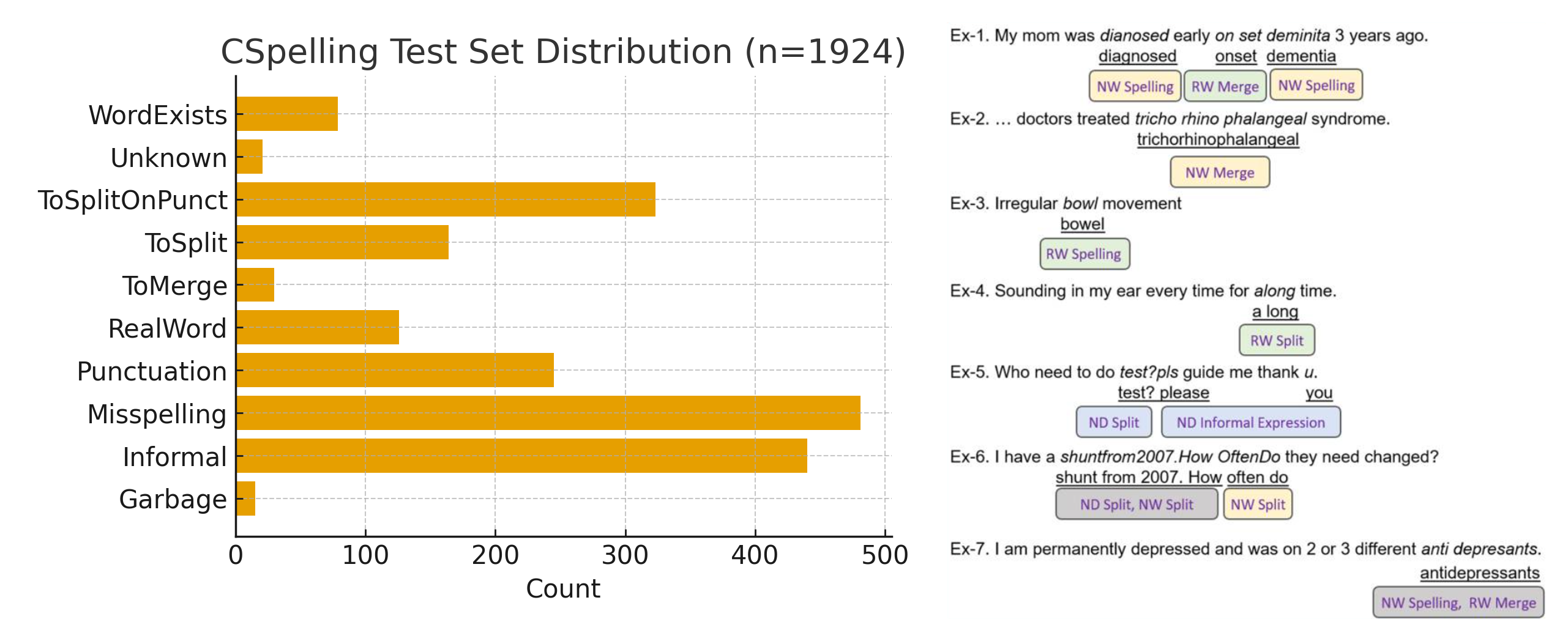} 
    \end{minipage}%
    \caption{ Consumer Health Cspelling Test Data Distribution. Both our methods and Gemini 2.5 pro are very good in detecting Misspelling error \textbf{(with c.a. 92\%)}, but very bad on Punctuation, Informality, and To-split/To Merge errors \textbf{(with c.a. 41\%)}.}
    \label{fig: Cspelling}
\end{figure*}

\begin{figure*}[ht]
    \centering
    \begin{minipage}{1\textwidth}
        \centering
        \includegraphics[width=1.1\textwidth]{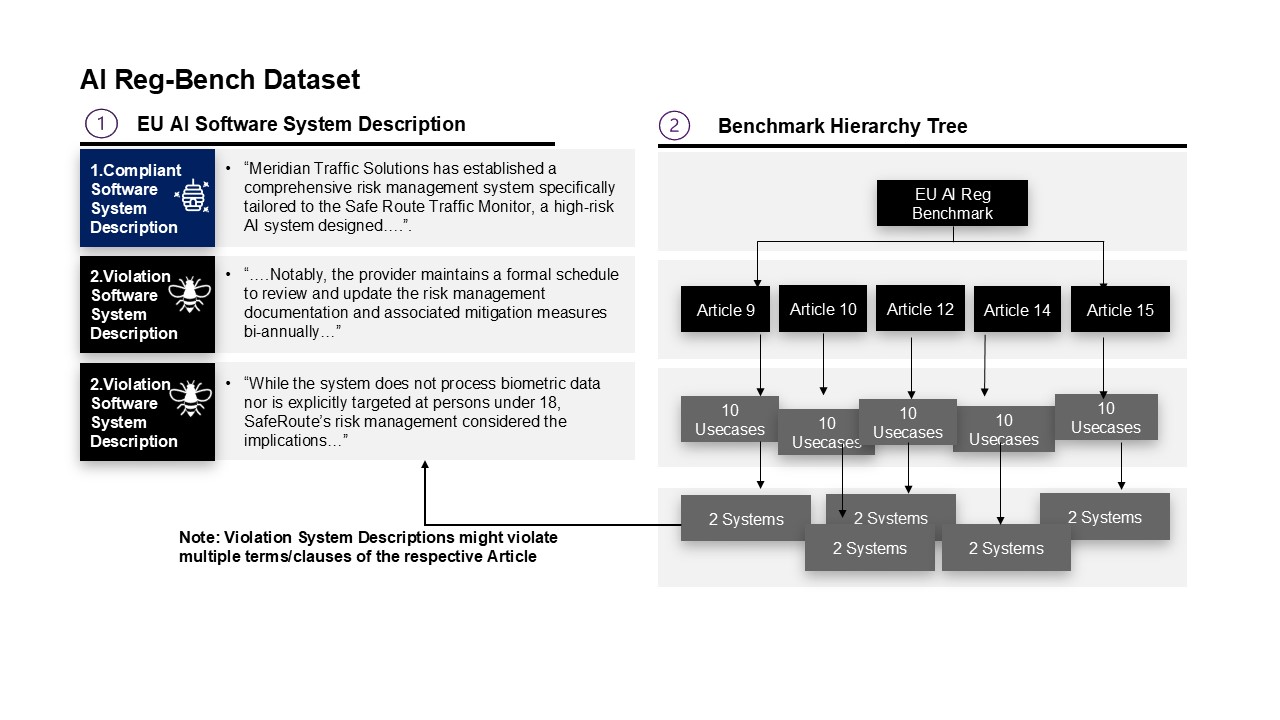} 
    \end{minipage}%
    \vspace{-50pt}
    \caption{AI Reg Benchmark Dataset.} 
    \label{fig: AIregbench}
\end{figure*}

\begin{figure*}[ht]
    \centering
    \begin{subfigure}{\textwidth}
        \centering
        \includegraphics[width=\linewidth]{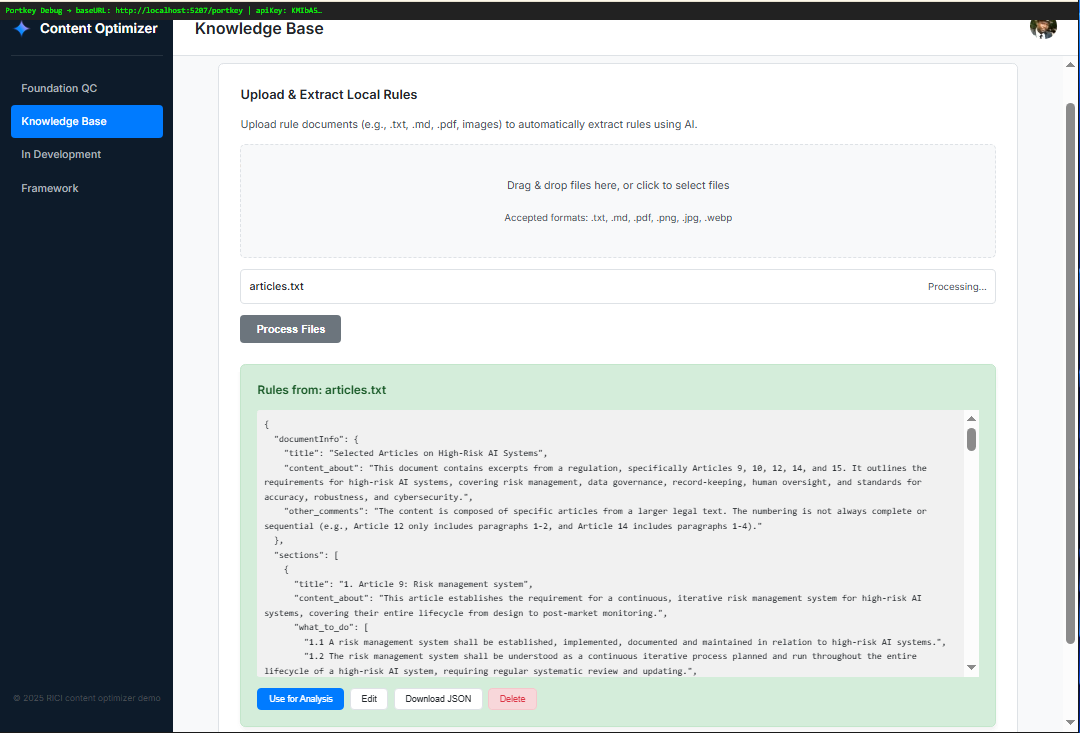}
        \caption{Rule Extraction Output}
        \label{fig:rule_extraction}
    \end{subfigure}
    \hfill
    \begin{subfigure}{\textwidth}
        \centering
        \includegraphics[width=\linewidth]{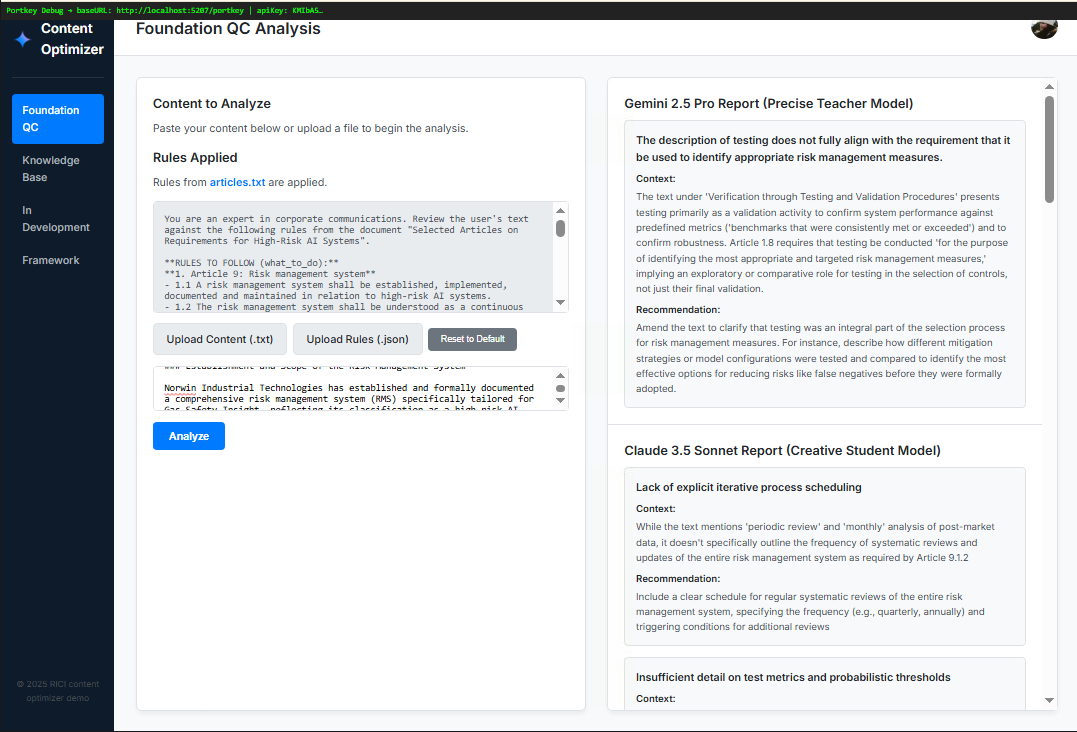}
        \caption{Verification Output (Student–Teacher Gemini/Claude)}
        \label{fig:verification}
    \end{subfigure}
\end{figure*}

In this section we provide the supplementary compiled together with the main paper includes:
\begin{itemize}

\item Ablation study on generalization, token count, latency, and
monetary cost of Student–Teacher dual-head LLM in Table~\ref {tab:cost1}, Table~\ref {tab:cost2} , Table~\ref {tab:cspelling_ablation}, Table~\ref{tab:ablationstudy1};

\item Property Dataset distribution on Table~\ref{table:roche-data}, and Consumer Health Cspelling and AI Reg-benchmark raw rata example on Figure~\ref{fig: Cspelling} and Figure~\ref{fig: AIregbench};

\item The Waterfall rule extraction and Student-Teacher prompt and rule verification prompt instruction lists in Table~\ref{tab:foundation_qc_rules}, Table~\ref{tab:rule_extraction_prompt}, Table~\ref{tab:verification_prompt}, and output example in Figure~\ref{fig:verification};

\end{itemize}

\subsection{Ablation study: generalization on Student-Teacher Model}

Here, we include an appendix ablation study using not only Gemini 2.5 Pro and Flash, but also Gemini Pro as teacher and a anthropic model as student models. This approach will allow us to determine the effectiveness of using models from the same family as student-teacher models vs a smarter proprietary model paired with a less intelligent open-source student model. It also demonstrates how well our dual-head framework generalizes as a plug-and-play mechanism for other LLM combinations.


The \textit{Claude-as-student} ablation study maintains high Recall but suffers substantial drops in Precision and Specificity, resulting in excessive false positives. This collapse in reliability significantly reduces overall Accuracy and F1-score, as confirmed in the ablation logs.

In contrast, the \textit{student--teacher baseline configuration} (Gemini Pro as teacher, same-family Gemini variant as student) achieves the strongest overall performance across metrics:
\begin{itemize}
    \item \textbf{Better Balance} ($\text{F1}=0.768$): achieves favorable trade-off between violation detection and false alarm suppression.
    \item \textbf{Exceptional Detection} ($\text{Recall}=0.95$): identifies nearly all true violations, which is critical for high-stakes regulatory compliance where missed violations must be minimized.
\end{itemize}

Conversely, the \textit{Gemini-Pro teacher + Claude student} variant demonstrates a high-risk, low-reward profile. Its degraded overall performance, reflected in a low F1-score ($\text{0.5568}$) and Accuracy ($\text{0.4245}$), is driven by a catastrophic lack of reliability. The Precision of only $0.4936$ and extremely low Specificity indicate that it fails to correctly identify compliant text and generates large numbers of false alarms, making it operationally impractical for compliance workflows.

\subsection{Ablation study: framework perform in terms of token count, latency, and monetary cost }
We also log token usage and per-request billing at the model level. In our pilot, calls to Gemini 2.5 Pro typically cost on the order of 3–8 ¢ (US cents) per request for ~2–11K tokens, while Gemini 2.5 Flash costs were sub-cent (e.g., ~0.4–0.6¢ for ~1.8–1.9K tokens). This enables tiered routing and industrial scale, where low-risk or preliminary checks use a lightweight flash model and ambiguous/high-risk cases are escalated to a teacher model, controlling the incremental overhead of multi-pass verification. (See Table~\ref {tab:cost1}, ~\ref {tab:cost2}).

In addition, we record per-request token usage and billing of production logs from different model providers. In our pilot, Gemini 2.5 Flash requests cost sub-cent (e.g., ~0.38–0.64¢ for ~1.8–1.9K tokens), while Gemini 2.5 Pro typically costs a few cents per request. For Anthropic endpoints, observed costs were also low (e.g., ~1.46–3.93¢ for ~4–11.9K tokens; ~0.32–0.37¢ per 1K tokens), which enables scalability for industry application \citep{explainableaibenchmark}.

\subsection{Ablation study: mathematical rationale of the Student–Teacher Dual Model Framework}
The Student–Teacher architecture is inspired by the practical knowledge-distillation paradigm \citep{hu2023teacherstudentarchitectureknowledgedistillation}. Classical teacher-student knowledge distillation is often formulated as minimizing the divergence between the teacher’s output distribution and the student’s output distribution, commonly through a KL-divergence or cross-entropy objective\citep{hu2023teacherstudentarchitectureknowledgedistillation}. 

In our setting, however, the goal is not model compression. The goal is controlled content optimization under compliance constraints. Therefore, we reinterpret the student-teacher mechanism as a dual-model \ref{fig: architecture} verification process: 1) The teacher model acts as a stable, high-accuracy evaluator, guides knowledge execution.
2) The student model acts as a more exploratory evaluator that can challenge, rediscover, or expand the teacher’s decision, it verifies commonly adopted rules and proposes conflicts or new ideas.
3) The human-in-the-loop layer resolves unresolved conflicts and the human reviewer resolves remaining disagreements (See our \href{https://demo-of-content-optimizer-338849523617.us-west1.run.app/}{Demo}).

\subsubsection{Formalizing as a Rule-Constrained Optimization Problem}

Let the input content be denoted as:
\[
x \in \mathcal{X},
\]
where \(x\) may represent a claim, paragraph, slide, document asset, or multi-modal content block.

Let the full enterprise compliance rule base be:
\[
\mathcal{R} = \{r_1, r_2, \ldots, r_N\},
\]
where each rule \(r_j\) belongs to our pre-defined quality-control dimensions:
\[
r_j \in \{L, R, B, T, C\}.
\]
Here, \(L\), \(R\), \(B\), \(T\), and \(C\) represent Language, Regulatory, Brand, Technical, and Content Structure, respectively.

The goal is to generate or refine the output \(x'\) that maximizes the utility of the content while minimizing the risk of quality and compliance:
\[
x^* =
\arg\max_{x'}
\left[
U(x') - \lambda V(x', \mathcal{R})
\right],
\]
where \(U(x')\) denotes the usefulness, clarity, or business value of the refined content, \(V(x', \mathcal{R})\) denotes the total violations under the applicable rule base, and \(\lambda\) is a risk-sensitivity coefficient.

In regulated pharmaceutical content generation, the Optimizer should prioritize reducing false negatives, especially for legal, regulatory, scientific, and medical accuracy checks, because the cost of missing a violation is usually higher than the cost of over-flagging a potential issue.

\subsubsection{Waterfall Filtering as Rule-Space Reduction}

Directly checking every input \(x\) against the complete rule base \(\mathcal{R}\) is inefficient and may increase hallucination risk due to excessive context length. If the full rule base contains \(N\) rules, exhaustive verification has complexity proportional to \(O(N)\).

To address this, Optimizer applies a waterfall filtering code function:
\[
\mathcal{R}_x = W(x, \mathcal{R}),
where:
\mathcal{R}_x \subset \mathcal{R}.
\]

The function \(W\) selects only the rules relevant to the specific input context. For example, the rule selection process follows a hierarchical sequence:
\[
\begin{aligned}
W:
&\text{IP}
\rightarrow
\text{Country}
\rightarrow
\text{Use Case} 
\rightarrow
\text{Topic} \\ &
\rightarrow
\text{Subtask (e.g., grammar, spelling, citation)}.
\end{aligned}
\]

This reduces the effective rule space from:
\[
|\mathcal{R}| = N, \quad
to \quad
|\mathcal{R}_x| = k,
\quad
k \ll N.
\]

Thus, Optimizer only evaluates the subset of rules that are relevant to the given content item. This has two benefits. First, it reduces computational cost and latency. Second, it reduces epistemic noise by preventing the LLM from reasoning over irrelevant or conflicting rules. This is particularly important in enterprise settings, where global, regional, therapeutic-area, brand, and channel-specific rules may overlap.

\subsubsection{ Agreement And Conflict}

The decisions from the teacher and student can be then partitioned into agreement and conflict sets.

The agreement set is defined as:
\[
\mathcal{A}
=
\left\{
r_j \in \mathcal{R}_x
:
\hat{y}^{T}_j = \hat{y}^{S}_j
\right\}.
\]

Rules in \(\mathcal{A}\) are treated as higher-confidence decisions because both models independently reach the same conclusion.

The conflict set is defined as:
\[
\mathcal{C}
=
\left\{
r_j \in \mathcal{R}_x
:
\hat{y}^{T}_j \neq \hat{y}^{S}_j
\right\}.
\]

Rules in \(\mathcal{C}\) represent uncertain or unstable cases. These cases require additional Teacher verification because the two models disagree on whether the rule has been violated.

The human review set is defined as:
\[
\mathcal{H} \subseteq \mathcal{C}.
\]

In practice, not all conflicts need to be escalated directly to human experts. Before human review, the conflict set \(\mathcal{C}\) is re-evaluated by the teacher model using a dedicated verification prompt. Specifically, the teacher model checks the input content \(x\) against the conflicting rules in \(\mathcal{C}\) and determines which conflicts are valid, invalid, or still uncertain. As a result, only the remaining unresolved or high-risk cases are passed to human experts. This means that human experts review only a small subset of the original rule space:
\[
|\mathcal{H}| \ll |\mathcal{R}_x| \ll |\mathcal{R}|.
\]

Manual review is transformed from a full-review problem into an exception-review problem.

\subsubsection{Conflict as a Loss Signal}

The disagreement between the teacher and student can be interpreted as a conflict loss:
\[
\mathcal{L}_{\text{conflict}}
=
\sum_{r_j \in \mathcal{R}_x}
\mathbb{1}
\left[
\hat{y}^{T}_j
\neq
\hat{y}^{S}_j
\right],
\]
where \(\mathbb{1}[\cdot]\) is an indicator function.

A high conflict loss indicates that rule interpretation is unstable. These unstable cases can be prioritized for additional verification, rerouting, or human review.
After human review, we obtain a human-approved label:
\[
y^{H}_j.
\]

The feedback loss can then be expressed as:
\[
\mathcal{L}_{\text{HITL}}
=
\sum_{r_j \in \mathcal{H}}
\left[
\ell(\hat{y}^{T}_j, y^{H}_j)
+
\ell(\hat{y}^{S}_j, y^{H}_j)
\right],
\]
where \(\ell(\cdot)\) represents classification loss, and facing task-specific disagreement penalty.

Each resolved conflict creates a reusable supervision tuple:
\[
(x, r_j, y^{H}_j).
\]

These tuples can be used to further improve future rule routing helping LLM for rule interpretation. In this sense, human-in-the-loop feedback does not merely correct a single output. It also provides structured learning signals for the LLM.

\subsubsection{How the Student--Teacher Mechanism Helps}

The Student--Teacher mechanism supports in three main ways.

\begin{enumerate}
    \item It improves rule adoption reliability. A rule decision is more trustworthy when two differently configured models independently reach the same conclusion.
    \item It improves recall for high-risk violations. Since the Student--Teacher system misses a violation only when both models fail to detect it, the dual-model design can reduce false negatives under the assumption of partially independent model errors.
    \item It improves traceability. Each rule-level decision can be linked to the input content, rule ID, teacher output, student output, agreement or conflict status, and human decision, e.g. reusable super vision tuple. This creates an auditable decision path.
\end{enumerate}

The teacher stabilizes rule interpretation, the student increases exploratory coverage, and the human-in-the-loop layer resolves unresolved conflicts. This provides a practical mathematical rationale for why the Student--Teacher can improve verifiability, reduce missed violations, and support scalable content optimization in regulated pharmaceutical workflows.

\begin{table}[t]
\centering
\small
\begin{tabular}{l l l l}
\hline
Provider/Model &  Sample Tokens & Cost & / 1K tok \\

\hline
Gemini 2.5 Pro   & 2.4K  & 3.39\textcent & 1.41\textcent \\
Gemini 2.5 Pro   & 5.0K  & 7.46\textcent & 1.49\textcent \\
Gemini 2.5 Flash & 1.9K  & 0.64\textcent & 0.34\textcent \\
Gemini 2.5 Flash & 1.8K  & 0.38\textcent & 0.21\textcent \\
Anthropic (Claude) & 11.0K & 3.51\textcent & 0.32\textcent \\
Anthropic (Claude) & 11.9K & 3.93\textcent & 0.33\textcent \\
Anthropic (Claude) & 4.0K  & 1.46\textcent & 0.37\textcent \\
\hline
\end{tabular}
\caption{Example per-request costs from system logs. Costs depend on provider/model tier, motivating cost-aware routing in multi-pass verification.}
\label{tab:cost1}
\end{table}

\begin{table}[t]
\centering
\begin{tabular}{l r}
\hline
Metric & Value \\
\hline
Total tokens Per week & 12.3M \\
Total requests & 2,000 \\
Latency (P50) & 2010 ms \\
Cost / request & \$0.02045 \\
Tokens / request & 6,150 \\
\hline
\end{tabular}
\caption{Pilot deployed cost and latency summary.}
\label{tab:cost2}
\end{table}

\begin{table*}[t]
\centering
\caption{Ablation study of the Student–Teacher framework using different LLMs as the teacher model. Mean accuracy and sample standard deviation are reported on the Consumer Health CSpelling ablation subset (Subset 1 and 5), where accuracy is defined as detected\,/\,ground-truth errors per card. The configuration using \textit{Gemini 2.5 Pro as teacher and Claude as student} shows improved detection of non-existent medical terms and misspellings, but exhibits weaker performance on punctuation and token-splitting errors. Nevertheless, it still substantially outperforms standalone Gemini 2.5 Pro.}

\begin{tabular}{lcc}
\hline
Model & Mean Accuracy (\%) & SD (\%) \\
\hline
Gemini Pro            & 15.1 & 21.7 \\
(Gemini 2.5 Pro  Teacher + Gemini 2.5 Flash Student)  & 40.1 & 26.4 \\
(Gemini 2.5 Pro  Teacher + Claude 3.5 Sonnet Student)      & 48.1 & 27.3 \\
\hline
\end{tabular}

\label{tab:cspelling_ablation}
\end{table*}

\begin{table*}[ht]
\centering
\caption{Ablation study of the Student–Teacher framework using different LLMs as the teacher model, and corresponding performance metrics on AIReg-Bench (N = 60 systems, Article 9) for regulatory compliance under the EU AI Act. Using models from the same family (a stronger model as teacher and a weaker but more diverse generator as student) yields better balance between detection and reliability. In contrast, the configuration with Gemini Pro as teacher and Claude as student produces substantially more false positives, reducing specificity without improving overall accuracy in violation detection.}

\label{tab:performance_comparison ab}
\begin{tabular}{l c c}
\toprule
\textbf{Metric} & \textbf{Student-Teacher Methods}  (\%) & \textbf{Student-Teacher Methods} \\
 & (Gemini 2.5 Pro as Teacher   &  (Gemini 2.5 Pro Teacher + \\
 & + Claude 3.5 Sonnet as Student)   &  + Gemini 2.5 Flash Student) \\
\midrule
Accuracy ($\frac{TP+TN}{Total}$) & 42.45 & \textbf{65.67} \\
Recall ($\frac{TP}{TP+FN}$) & 63.89 & \textbf{95.00} \\
Precision ($\frac{TP}{TP+FP}$) & 49.36 & \textbf{64.41} \\
F1-Score  & 55.68 & \textbf{76.86} \\
Specificity ($\frac{TN}{TN+FP}$) & 14.49 & \textbf{60.0} \\
\bottomrule
\end{tabular}
\label{tab:ablationstudy1}
\end{table*}

\begin{table*}[ht]
    \centering
    \caption{Distribution of Property Audio and Video Data Across Medical Diseases Specialties.}
    \renewcommand{\arraystretch}{1.2}
    \rowcolors{2}{gray!15}{white}
    \begin{tabular}{l|c|c||l|c|c}
        \toprule
        \textbf{Specialty} & \textbf{Audio} & \textbf{Video} & \textbf{Specialty} & \textbf{Audio} & \textbf{Video} \\
        \midrule
        Oncology & 208 & 8934 & Ophthalmology & 159 & 2862 \\
        Sample setiovascular & 1 & 14 & Respiratory Disease & 16 & 467 \\
        Dermatology & 0 & 30 & Nephrology & 1 & 380 \\
        Hematology & 67 & 3606 & Not Applicable & 59 & 2853 \\
        Immunology & 144 & 510 & Movement Disorder & 9 & 289 \\
        Infectious Disease & 1 & 239 & Inflammatory Disease & 20 & 222 \\
        Metabolism & 0 & 6 & Neuroscience & 202 & 4914 \\
        \bottomrule
    \end{tabular}
    \label{table:roche-data}
\end{table*}

\begin{table*}[ht]
\centering
\caption{Foundation QC Rules Prompt (System Instruction e.g.  Global Copy Style Guide Book as default)}
\begin{tabular}{p{0.20\linewidth} p{0.75\linewidth}}
\toprule
\rowcolor{rocheblue!20}\textbf{System Prompt} & \textbf{Content} \\
\midrule

\textbf{System Role} &
You are an \textcolor{rocheblue}{expert in corporate communications and compliance}. Review the user's text against the following Foundation QC (Quality Control) rules. Using Default Global Copy Style Guide Book. Focus on UK English spelling and grammar unless otherwise specified.\\

\midrule
\textbf{Spelling Rules -- \textcolor{rocheblue}{What to Do}} &
\begin{itemize}
\item 1.1 Use UK English for all corporate communications.
\item 1.2 For non-commercial or internal use, generally follow UK spelling, unless content is purely for a US audience or the brief requires US spelling.
\item 1.3 Use ``analogue'' when meaning ``something similar to something else'' or referring to chemical compounds.
\item 1.4 Check the BNF (medicines.org.uk/emc) for correct spelling of UK proprietary and generic drug names.
\textbf{......}
\end{itemize}
\\

\midrule
\textbf{Spelling Rules -- \textcolor{rocheblue}{What to Prohibit}} &
\begin{itemize}
\item 2.1 Avoid US spellings in UK English such as acknowledgment, etiology, aging, anemia, anesthetic, analog, behavior, cesarean, center, coloration, corticotropin, dialog (except for UI), diarrhea, dyspnea, enroll, favor,  ... practice (verb), program (unless IT), signaling, sulfur, tumor.
\item 2.2 Do not add spaces around ampersands.
\item 2.3 Do not use -t where -ed is preferred (learnt, spelt, burnt).
\item 2.4 Do not use -ize in UK English except defined exceptions.
\textbf{......}
\end{itemize}
\\

\midrule
\textbf{\textcolor{rocheblue}{Other Comments}} &
 Officially uses UK English for all corporate communications. Exceptions apply for US audiences or specific briefs. The rulebook contains UK vs US spellings, ampersand usage, drug naming standards, past tense guidance, and rules for -ise/-ize. \\

\midrule
\textbf{Analysis Requirements (Chain-of-Thought)} &
\begin{enumerate}
\item Carefully examine the plans against each relevant rule in the rule books
\item Identify any areas where the plans may not comply with the rules
\item \textcolor{rocheblue}{Cite specific rules} when discussing compliance issues
\item Consider both explicit requirements and implicit principles in the rule books
\end{enumerate}
\\

\midrule
\textbf{Output Format} &
Respond with a JSON object. The object should have a single key "issues" which is an array of objects. 
Each violation must include:
\begin{enumerate}
\item \texttt{\textcolor{rocheblue}{"issue"}} rule ID \textbf{(MUST HAVE)} for the rule that was violated
\item \texttt{\textcolor{rocheblue}{"context"}} The exact text snippet from the user's content where the violation occurred
\item \texttt{\textcolor{rocheblue}{"recommendation"}} A clear suggestion on how to fix the violation.
\end{enumerate}
If no issues are found, return \texttt{"issues": []}. \\
\bottomrule
\end{tabular}
\label{tab:foundation_qc_rules}
\end{table*}

\begin{table*}[ht]
\centering
\caption{Rule Extraction Prompt (System Instruction for JSON-Structured Regulatory Rule Extraction and Parsing)}
\begin{tabular}{p{0.20\linewidth} p{0.75\linewidth}}
\toprule
\rowcolor{rocheblue!20}\textbf{System Prompt} & \textbf{Content} \\
\midrule

\textbf{System Role} &
You are a compliance and regulation expert. I will provide you with content from a document (text or an image). Your task is to carefully analyze the document and extract all the rules into a JSON file format. \\

\midrule
\textbf{Task Overview} &
The JSON object should include the document's title, a summary of its content, and any other general comments.  
It must also contain a \texttt{sections} array, where each section details a specific set of rules, including what to do, what to prohibit, and any other relevant comments. \\

\midrule
\textbf{Critical Instruction} &
\textbf{When extracting rules for \texttt{what\_to\_do} and \texttt{what\_to\_prohibit}, you MUST use the exact wording, phrasing, and sentences from the original document. Do not rephrase, summarize, or generate new text for the rules. Copy them verbatim from the source.} \\

\midrule
\textbf{Required JSON Structure} &
\begin{verbatim}
{
  "documentInfo": {
    "title": "Document Title Here",
    "content_about": "A brief description of this document.",
    "other_comments": "Any other general comments about the document."
  },
  "sections": [
    {
      "title": "Title of the Rule Section",
      "content_about": "Description of this rule section.",
      "what_to_do": ["Rule 1 of what to do.", "Rule 2..."],
      "what_to_prohibit": ["Rule 1 of what to avoid.", "Rule 2..."],
      "other_comments": "Optional comments for this specific section."
    }
  ]
}
\end{verbatim}
\\

\midrule
\textbf{Final Instruction} &
\texttt{Now, analyze the following content and provide the structured JSON output.} \\

\bottomrule
\end{tabular}
\label{tab:rule_extraction_prompt}
\end{table*}

\begin{table*}[ht]
\centering
\caption{Verification Prompt (Teacher Model)}
\begin{tabular}{p{0.20\linewidth} p{0.75\linewidth}}
\toprule
\rowcolor{rocheblue!20}\textbf{Prompt} & \textbf{Content} \\
\midrule

\textbf{Task Description} &
Verification Task using a teacher model to perform authoritative final review of previously flagged issues.\\

\midrule
\textbf{Original Text} &
\texttt{\$\{userText\}} \\

\midrule
\textbf{Original Rules} &
\texttt{\$\{currentSystemInstruction\}} \\

\midrule
\textbf{Issues to Re-evaluate} &
The following issues were identified by different models and require a final, authoritative check.

\begin{verbatim}
${issuesToVerify.map(i => `- Rule: ${i.issue}
${ Context: "${i.context}"`).join('\n')}
\end{verbatim}
\\

\midrule
\textcolor{rocheblue}{\textbf{Teacher Model Instructions}} &
\begin{enumerate}
\item Critically re-evaluate each issue against the original text and rules.
\item If multiple issues point to the same underlying error (e.g., a single typo violating two different spelling rules), consolidate them into a single issue in your response.
\item For each final issue, determine if it is a valid violation.
\end{enumerate}
\\

\midrule
\textbf{Output Format Requirements} &
Respond with a JSON object. The object must have a single key \texttt{"issues"} which is an array of objects. Each object represents one of the re-evaluated issues and must have the following FOUR keys:

\begin{enumerate}
\item \texttt{"issue"}: The most appropriate rule ID for the violation.
\item \texttt{"context"}: The exact snippet from the user text.
\item \texttt{"recommendation"}:  Your final verdict and justification. If it is NOT a valid violation, explain why it was likely flagged incorrectly.
\item \texttt{"isValid"}:  A boolean value. Set to 'true' if the issue is a valid violation, and 'false' otherwise.
\end{enumerate}
\\

\bottomrule
\end{tabular}
\label{tab:verification_prompt}
\end{table*}

\clearpage

\end{document}